\documentclass{article}

\usepackage[final,nonatbib]{nips_2016}
\usepackage[utf8]{inputenc} 
\usepackage[T1]{fontenc}    
\usepackage{url}            
\usepackage{booktabs}       
\usepackage{amsfonts}       
\usepackage{nicefrac}       
\usepackage{microtype}      
\usepackage{graphicx}
\usepackage{amsmath}		
\usepackage{tikz}			
\usepackage{todonotes}
\usepackage{mathabx}
\usetikzlibrary{matrix, positioning}
\usepackage[numbers]{natbib}

\definecolor{verde}{RGB}{16,150,24}
\definecolor{viola}{RGB}{153,0,153}
\definecolor{giallo}{RGB}{255,153,0}
\definecolor{rosso}{RGB}{220,57,18}

\newcommand\colA[1]{{\color{rosso} #1}}

\makeatletter
\def\@xfootnote[#1]{%
  \protected@xdef\@thefnmark{#1}%
  \@footnotemark\@footnotetext}
\makeatother

\title{Neural Document Embeddings for Intensive Care Patient Mortality Prediction}

\author{
  Paulina Grnarova\footnote[\!*]{}\ , Florian Schmidt\footnote[\!*]{}\ , Stephanie L.\ Hyland and Carsten Eickhoff\\
  Department of Computer Science\\
  ETH Zurich\\
  Zurich, Switzerland\\
  \texttt{firstname.lastname@inf.ethz.ch}\\
  \footnotesize{* contributed equally}
}

\begin{document}

\maketitle
\vspace{-0.5cm}
\begin{abstract}
We present an automatic mortality prediction scheme based on the unstructured textual content of clinical notes. Proposing a convolutional document embedding approach, our empirical investigation using the MIMIC-III intensive care database shows significant performance gains compared to previously employed methods such as latent topic distributions or generic doc2vec embeddings. These improvements are especially pronounced for the difficult problem of post-discharge mortality prediction.
\end{abstract}

\section{Introduction}
The steadily growing amount of digitized clinical data such as health records, scholarly medical literature, systematic reviews of substances and procedures, or descriptions of clinical trials holds significant potential for exploitation by automatic inference and data mining techniques. Besides the wide range of clinical research questions such as drug-to-drug interactions~\cite{wienkers2005predicting} or quantitative population studies of disease properties~\cite{wren2005data}, there is a rich potential for applying data-driven methods in daily clinical practice for key tasks such as decision support~\cite{kawamoto2005improving} or patient mortality prediction~\cite{moreno2005saps}. The latter task is especially important in clinical practice when prioritizing allocation of scarce resources or determining the frequency and intensity of post-discharge care.

There has been an active line of work towards establishing probabilistic estimators of patient mortality both in the clinical institution as well as after discharge~\cite{pirracchio2015mortality,che2016recurrent,johnson2016machine}. The authors report solid performance on both publicly available and proprietary clinical datasets.

In spite of these encouraging findings, we note that most competitive approaches rely on time series and demographic information while algorithmic processing of the unstructured textual portion of clinical notes remains an important, yet, to date, insufficiently studied problem. The few existing advances towards tapping into this rich source of information rely on term-wise representations such as tf-idf embeddings~\cite{ghassemi2014unfolding} or distributions across latent topic spaces~\cite{lehman2012risk}. 

This intuitively appears sub-optimal since several studies have independently highlighted the importance of accounting for phrase compositionality manifested, \textit{e.g.}, in the form of negations~\cite{kuhn2016implicit}, or long-range dependencies in clinical resources. Models that solely rely on point estimates of term semantics cannot be assumed to adequately capture such interactions. 

In this paper, we aim to address these shortcomings by presenting a convolutional neural network architecture that explicitly represents not just individual terms but also entire phrases or documents in a way that preserves such subtleties of natural language.

The remainder of this paper is structured as follows: Section~\ref{sec:model} introduces our model and our objective function. Subsequently, in Section~\ref{sec:experiments}, we empirically evaluate the model against two competitive baselines on the task of intensive care unit (ICU) mortality prediction on the popular MIMIC-III database~\cite{johnson2016mimic}. Finally, Section~\ref{sec:conclusion} concludes with a brief discussion of our findings. 

\section{Model}
\label{sec:model}
While simple feed-forward architectures, such as the doc2vec scheme~\cite{le2014distributed}, have been established as versatile plug-in modules in many machine learning applications~\cite{lee2016sentiment,lau2016empirical}, they are inherently incapable of directly recognizing complex multi-word or multi-sentence patterns. However, constructions such as \textit{no sign of pneumothorax} are frequently encountered in clinical notes and encode crucial information for the task of mortality prediction.

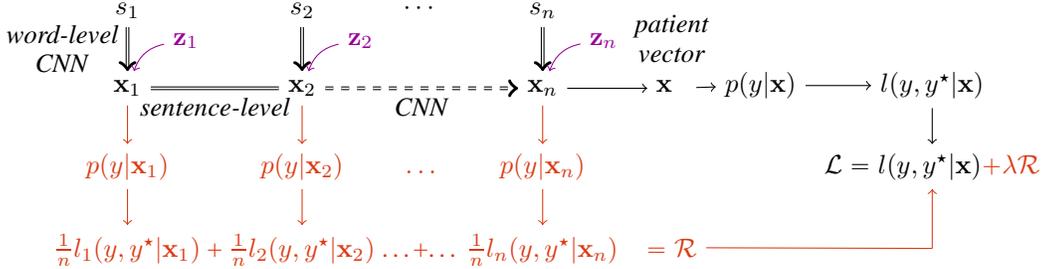
\begin{figure}[ht]
	\begin{tikzpicture}[scale=0.5]
	\matrix (m) [matrix of math nodes,row sep=1.4em,column sep=0.4em,minimum width=0.1em]
  {
     s_1 & s_2 & \cdots & s_n & & &\\
     \mathbf x_1 & \mathbf x_2 &  & \mathbf x_n & \mathbf x_{\phantom{n}}& p( y|\mathbf x) & l(y,y^\star|\mathbf x)\\
     \colA{p( y|\mathbf x_1)} & \colA{p( y|\mathbf x_2)} &\colA{\dots} & \colA{p( y|\mathbf x_n)} & & & \mathcal L = l( y, y^\star|\mathbf x) \colA{+ \lambda\mathcal R} \\
      \colA{\frac 1 n l_1(y,y^\star|\mathbf x_1)} & \colA{\frac 1 n l_2(y,y^\star|\mathbf x_2)} &  & \colA{\frac 1 n l_n(y,y^\star|\mathbf x_n)} & \colA{=\mathcal R}& {}& {\phantom{\cdot}}\\
  };
  \path[->] (m-1-1) edge [double] node [left, align=center] {\textit{word-level}\\ \textit{CNN}} (m-2-1);
  \path[->] (m-1-2) edge [double] node [left] {} (m-2-2);
  \path[->] (m-1-4) edge [double] node [left] {} (m-2-4);

  \draw[->, dashed] (m-2-2) edge[double] node [below] { \textit{CNN}}  (m-2-4);
  \draw[-] (m-2-1) edge[double] node [below] {\textit{sentence-level}} (m-2-2);
  \draw[->] (m-2-4)  -- (m-2-5);
  \draw[->] (m-2-5)  -- (m-2-6);
  \draw[->] (m-2-6)  -- (m-2-7);

  \draw[->,color=rosso] (m-2-1)  -- (m-3-1);
  \draw[->,color=rosso] (m-2-2)  -- (m-3-2);
  \draw[->,color=rosso] (m-2-4)  -- (m-3-4);

  \draw[->,color=rosso] (m-3-1)  -- (m-4-1);
  \draw[->,color=rosso] (m-3-2)  -- (m-4-2);
  \draw[->,color=rosso] (m-3-4)  -- (m-4-4);

   \path[opacity=0] (m-4-1) edge node [midway, opacity=1, color=rosso] {+} (m-4-2);
   \path[opacity=0] (m-4-2) edge node [midway, opacity=1, color=rosso] {\dots+\dots} (m-4-4);
  \coordinate[right=of m-4-5, below= of m-3-7, yshift=0.6em]  (corner) ;
  \draw[color=rosso] (m-4-5)  -- (corner);
  \draw[->,color=rosso] (corner)  -- (m-3-7);

  \draw[->] (m-2-7)  -- (m-3-7);

  \node [above = 0 em of m-2-5, align=center] {\textit{patient} \\ \textit{vector}};

  \node [above right = 0.5 em and 0.5 em of m-2-1, color=viola] (i-1) {$\mathbf z_1$};
  \draw[->, bend right, color=viola] (i-1) edge (m-2-1);
  \node [above right = 0.5 em and 0.5 em of m-2-2, color=viola] (i-1) {$\mathbf z_2$};
  \draw[->, bend right, color=viola] (i-1) edge (m-2-2);
  \node [above right = 0.5 em and 0.5 em of m-2-4, color=viola] (i-1) {$\mathbf z_n$};
  \draw[->, bend right, color=viola] (i-1) edge (m-2-4);

	\end{tikzpicture}
    \caption{Model architecture: In black, our basic architecture. In red, target replication. In violet, optional note information $z_i$ introduced in the next section. The CNN layers are depicted by double arrows. For clarity, we omit the word-vectors that serve as input to the initial CNN.}
    \label{figArchitecture}
\end{figure}

Following recent work in document classification~\cite{yang2016} and dialogue systems~\cite{serbanSBCP15}, we adopt a two-layer architecture. Let $d=\langle s_1,\dots,s_n\rangle$ denote a patient's record comprising $n$ sentences. Our first layer independently maps sentences $s_i$ to sentence vectors $\mathbf x_i\in\mathbb R^{D_S}$. The second layer combines $\langle \mathbf x_1,\dots,\mathbf x_n\rangle$ into a single patient representation $\mathbf x\in\mathbb R^{D_P}$. For both levels we use convolutional neural networks (CNNs) with max-pooling which have shown excellent results on binary text classification tasks \cite{kim2014}, \cite{severyn2015}.    Following work by Severyn \textit{et al}~\cite{severyn2015}, we use word-embeddings to provide vector-input for the first CNN layer. Finally, the output of our model is $p(y),  y\in[0,1]$, the estimated mortality probability, and our objective is the cross entropy $l(y, y^\star)$ where $y^\star$ is the ground-truth label. The graph rendered in black in Figure \ref{figArchitecture} depicts this basic architecture. 
\paragraph{Target replication} The performance of the basic model presented above is promising but not yet satisfying. For similar long-sequence prediction problems,~\cite{lipton15replication} and~\cite{dai2015} have noted that it is beneficial to replicate the loss at intermediate steps. Following their approach, we compute an individual softmax mortality probability $p_i(y|\mathbf x_i)$ for every sentence $i=1\dots n$ and incorporate $n$ additional cross entropy terms into our final objective. For a corpus $\mathcal D$ containing patients $d_1,\dots d_N$ and corresponding labels $y^\star_1,\dots d^\star_N$ we seek to minimize:
\begin{align}
    \mathcal L &=\sum_{(d^{(j)},{y^\star}^{(j)})\in\mathcal D}\mathcal L(d^{(j)},{y^\star}^{(j)})\\
    \mathcal L (d=\langle s_1,\dots,s_n\rangle,y^\star)&= l(y, {y^\star}|\mathbf x) + \lambda\mathcal R = l(y, {y^\star}|\mathbf x) + \frac{\lambda}{n}\sum_{i=1}^nl_i(y, y^\star|\mathbf x_i) 
\end{align}
$\mathcal R$ can be interpreted as the average prediction error at the sentences level, effectively bringing the classification loss closer to the word-level and regularizing the first CNN to learn sentence representations tailored to the mortality prediction problem. The hyper-parameter $\lambda$ determines the strength of the regularizer.
\paragraph{Incorporating note information} End-to-end neural network architectures such as ours allow for easy incorporation of additional information that can increase predictive power. Every note in our collection has a \emph{category} associated such as \textit{nursing}, \textit{physician} or \textit{social work}. Providing this information to our classifier can help to reliably assess the importance of individual sentences for the classification task. To exploit this information, we embed all 14 categories into a vector space $\mathbb R^{D_C}$ and concatenate every sentence vector $\mathbf x_i$ with its associated category vector $\mathbf z_i$. 
\section{Experiments}
\label{sec:experiments}

We evaluate the proposed method on three standardized ICU mortality prediction tasks. On the basis of a patient's electronic health record, we predict whether the patient will die (1) during the hospital stay, (2) within 30 days after discharge, or, (3) within 1 year after discharge, and report AUC as an evaluation measure.



\subsection{Data}
\label{sub:data}
MIMIC-III~\cite{johnson2016mimic} is an openly-accessible critical care database, comprising 46,520 patients with 58,976 hospital stays. It contains measurements of patient state (through vital sign, lab tests and other variables) as well as procedures and treatments. Crucially, it also contains over 2 million unstructured textual notes written by healthcare providers.

Following the data filtering and pre-processing steps in~\cite{ghassemi2014unfolding}, we restrict to adults ($\geq$18 years old) with only one hospital admission. Most importantly, we exclude notes from the \emph{discharge summary} category and any notes recorded after the patient was discharged. This results in 31,244 patients with 812,158 notes. 13.82\% of patients died in the hospital, 3.70\% were discharged and died within thirty days, and 12.06\% were discharged and died within a year. We randomly sample 10\% of the patients for the test set, and 10\% for the validation set. The remaining 80\% of the patients are used during training. We construct the vocabulary by keeping the 300K most frequent words across all notes and replace all the words which are not part of the vocabulary with an out-of-vocabulary token.

\subsection{Baselines}
\label{sub:baseline}

\paragraph{LDA based model}
We recreate the LDA-based Retrospective Topic Model from ~\cite{ghassemi2014unfolding}. This model is the state-of-the-art method for mortality prediction on unstructured data from MIMIC II. We recreate the model on MIMIC III, and closely follow their preprocessing and hyperparameter settings. We tokenize each note and remove all stopwords using the Onix stopword list \footnote{\url{www.lextek.com/manuals/onix}}. The vocabulary is constructed as the union of the 500 most informative words in each patient's note based on a tf-idf metric. All words which are not part of the vocabulary are removed. We keep the number of topics to be 50 and set the LDA priors for the topic distributions and the topic-word distributions to $\alpha = \frac{50}{numberTopics}$ and $\beta = \frac{200}{vocabularySize}$, respectively. We train a separate linear kernel SVM on the per-note topic distributions to predict the mortality for each task. 

Since SVM classifiers are sensitive to significant class-imbalances, we follow~\cite{ghassemi2014unfolding} in randomly sub-sampling the patients who did not die in the training sets to reach a ratio of 70\%/30\% between the negative and positive class. We do not modify the distribution of classes within the test and validation set. The LDA vectors are trained on the entire training data, but the SVM classifiers are trained using the vectors from the down-sampled training sets only. 

\paragraph{Feed-forward Neural Network}
\label{sub:d2v}
As our second baseline we use the popular distributed bag of words (DBOW) scheme proposed by Le and Mikolov~\cite{le2014distributed}. In a range of initial experiments, we determined the DBOW architecture (rather than the distributed memory alternative) and an embedding space dimensionality of $400$ to be optimal in terms of accuracy and generality. Using the same pre-processing as for the LDA baseline, we train separate linear SVMs for each task.

\subsection{Parameters and Pretraining}
\label{sub:parameters}
We pre-train 50-dimensional word vectors on the training data using the word2vec implementation of the gensim~\cite{gensim} toolbox. Our word-level CNN uses 50 filters of sizes 3,\ 4 and 5 resulting in a sentence representation of size $D_S=150$. We embed categories in $D_C=10$ dimensional space and use 50 filters of size 3 for the sentence-level CNN resulting in a patient representation of size $D_P=50$. Furthermore, we regularize the fully connected layer before our final softmax by l2-regularization on the weights and dropout with keep probability 0.8. 

\subsection{Results}
\label{sub:results}


Table~\ref{tab:results} summarizes the results of the three models on all tasks. Across all methods there seems to be a general tendency that labels further in the future are harder to predict. We observe that both neural models are superior to the LDA baseline, in particular on the two harder tasks. Furthermore, our two-level CNN model outperforms doc2vec by a significant margin on all tasks.

\begin{table}[ht]
  \caption{MIMIC-III Mortality Prediction AUC}
  \label{tab:results}
  \centering
  \begin{tabular}{llll}
    \toprule
    Task     & LDA 	& doc2vec 	& CNN \\
    \midrule
    Hospital & 0.930			 			& 0.930 & \textbf{0.963}\\
    30-day   & 0.800		 				& 0.831	& \textbf{0.858}\\
    1-year   & 0.790		   				& 0.824 & \textbf{0.853}\\
    \bottomrule
  \end{tabular}
\end{table}

To highlight the effectiveness of the target replication, Table \ref{tab:CNNresults} shows the results of our model with and without target replication. We report on 30-days post-discharge, but performance on the other tasks is comparable. 

\begin{table}[ht]
  \caption{Performance analysis for target replication}
  \label{tab:CNNresults}
  \centering
  \begin{tabular}{lllll}
    \toprule
    Model     & without target replication & with target replication \\
    \midrule
    AUC & 		 		0.682 & 0.858\\
    \bottomrule
  \end{tabular}
\end{table}

The results of our CNN show that modeling sentence and document structure explicitly results in noticeable performance gains. In addition, learning sentence representations and training them in our regularizer on the classification task, enables us to retrieve a patient's most informative sentences. This allows an inspection of the model's features, similar to LDA's topic distributions but on the sentence level. This stands in stark contrast to doc2vec's generic document representations. To showcase these features, Table~\ref{tab:sentences} shows a patient's top five sentences indicating likelihoods of survival and death respectively.

\begin{table}[ht]
\caption{The three highest and three lowest scoring sentences of one patient in the 1-year task.}
  \begin{tabular}{ll}
    \toprule
    	P(survival) high& the remaining support lines are unchanged .\\
      & no effusion .\\
      &the cardiomediastinal contours are normal .\\
       \midrule
      P(survival) low&now found to have metastatic lesions in her brain .\\
      &impression UNK multiple large enhancing masses within the brain with \\
      &\phantom{bla}\qquad\ $\hookrightarrow$ surrounding vasogenic edema most consistent with .\\
      &enhancing lesions in the right temporal lobe and right mid brain consistent\\
      &\phantom{bla}\qquad $\hookrightarrow$ with metastatic disease .\\
    \bottomrule
  \end{tabular}
  
  \label{tab:sentences}
\end{table}
While most patients' top-scoring sentences look promising, a careful study of the predictions reveals that some neutral sentences can be ranked too highly in either direction. This is due to the model's inability to appropriately handle sentences that do not help to distinguish the two classes. We plan to address this in the future by a more advanced attention mechanism.

\section{Conclusion}
\label{sec:conclusion}
In this paper we developed a two-layer convolutional neural network for the problem of ICU mortality prediction. On the MIMIC-III critical care database our model outperforms both existing BOW approaches and the popular doc2vec neural document embedding technique on all three tasks. We conclude that accounting for word and phrase compositionality is crucial for identifying important text patterns. Such findings have impact beyond the immediate context of automatic prediction tasks and suggest promising directions for clinical machine learning research to reduce patient mortality.  



\small
\bibliographystyle{plain}
\bibliography{ref.bib}

\end{document}